\documentclass[11pt]{article}

\usepackage[final]{acl}

\usepackage{times}
\usepackage{latexsym}
\usepackage{multirow}
\usepackage{subcaption} 
\usepackage{amsmath}
\usepackage[T1]{fontenc}

\usepackage[utf8]{inputenc}
\usepackage{booktabs}

\usepackage{tabularx}
\usepackage{makecell}
\usepackage{microtype}

\usepackage{inconsolata}

\usepackage{graphicx}

\title{Fine-tuning with Hierarchical Prompting for Robust Propaganda Classification Across Annotation Schemas}

\author{
\textbf{Lukas Stähelin}\textsuperscript{1}\thanks{Equal contribution},\textbf{Veronika Solopova}\textsuperscript{1,2}\footnotemark[1],
\textbf{Max Upravitelev}\textsuperscript{1,2},  
\textbf{David Kaplan}\textsuperscript{1},
\textbf{Ariana Sahitaj}\textsuperscript{1,2} \AND
\textbf{Premtim Sahitaj}\textsuperscript{1,2},
\textbf{Charlott Jakob}\textsuperscript{1,2},
\textbf{Sebastian Möller}\textsuperscript{1,2},
\textbf{Vera Schmitt}\textsuperscript{1,2,3} \quad \\\\
\textsuperscript{1}Technische Universität Berlin, QU Lab, XplaiNLP Group, Berlin, Germany \\
\textsuperscript{2}German Research Center for Artificial Intelligence (DFKI), Berlin, Germany \\
\textsuperscript{3}Centre for European Research in Trusted AI (CERTAIN) \\
{\small \textbf{Correspondence:} \texttt{veronika.solopova@tu-berlin.de}}
}

\begin{document}
\maketitle
\vspace{0.8em}
\begin{abstract}
Propaganda detection in social media is challenging due to noisy, short texts and low annotation agreements. We introduce a new intent-focused taxonomy of propaganda techniques and compare it against an established, higher-agreement schema. Along three dimensions (model portfolio, schema effects, and prompting strategy) we evaluate the taxonomies as a classification task with the help of four language models (GPT-4.1-nano, Phi-4 14B, Qwen2.5-14B, Qwen3-14B). Our results show that fine-tuning is essential, since it transforms weak zero-shot baselines into competitive systems and reveals methodological differences that are hidden using base models. Across schemas, the Qwen models achieve the strongest overall performance, and Phi-4 14B consistently outperforms GPT-4.1-nano. Our hierarchical prompting method (\textbf{HiPP}), which predicts fine-grained techniques before aggregating them, is especially beneficial after fine-tuning and on the more ambiguous, low-agreement taxonomy, while remaining competitive on the simpler schema. The HQP dataset, annotated with the new intent-based labels, provides a richer lens on propaganda’s strategic goals and a challenging benchmark for future work on robust, real-world detection.
\end{abstract}

\section{Introduction}
Propaganda has long been used to shape public opinion and influence political outcomes \cite{jowett2018propaganda}, while online platforms amplify its reach. The Russia–Ukraine war exemplifies this evolution, with propaganda deployed as part of hybrid warfare \cite{perez2022strategic, zhdanova2017computational}. Russian campaigns manipulate perceptions, sow distrust, and polarize societies, making automatic detection of propaganda vital for safeguarding democratic processes \cite{wardle2017information, bayer2021disinformation}.

While propaganda detection has been explored in long-form news articles \cite{martino2020semeval}, short-form social media introduces additional challenges: limited context, informal language, and heavy use of abbreviations \cite{vijayaraghavan2022tweetspin, Maarouf_2024}. Subtle linguistic cues and context-dependent meanings often lead to low inter-annotator agreement \cite{srba2024survey, hasanain-etal-2024-gpt}. Propaganda also evolves over time \cite{solopova-etal-2023-evolution, solopova-etal-2024-check}, with AI-generated content becoming increasingly prevalent on social media \cite{doi:10.1142/9789811281860_0009}. These challenges are amplified by trade-offs in annotation schema design. Realistic schemas capture the diversity of real-world propaganda but reduce annotation reliability, while simplified ones boost agreement and learnability at the cost of ecological realism. Understanding how methodological choices interact with schema design is essential for building robust detection systems. Existing taxonomies, such as the span-level schemas by \citet{da-san-martino-etal-2020-semeval, sprenkamp2023large} and the recent hybrid hierarchy of \citet{sahitaj-etal-2025-hybrid}, are all technique-centric, enumerating persuasive rhetorical devices (e.g., loaded language, appeals to fear). Our new schema instead groups surface techniques by communicative intent, introducing high-level categories such as \textit{Shift Blame and Justify Aggression}, \textit{Manufacture Consent and Identity}, and \textit{Confuse and Distract}. These labels capture strategic goals like reframing moral responsibility, building in-group loyalty, or generating uncertainty. Prior schemas either scatter these constructs across multiple tags or do not explicitly encode. By reorganising techniques around these intents, our taxonomy is conceptually closer to framing-style analyses and provides a more explanatory lens on why a message is propagandistic, although its empirical advantages beyond the current dataset remain to be validated.

In addition, we propose a framework for analyzing LLM-based propaganda classification along three interacting dimensions: (i) annotation schema design and its associated reliability, (ii) supervision regime (zero-shot vs. fine-tuned), and (iii) prompting strategy (direct vs. hierarchical). Importantly, we treat these three dimensions as orthogonal factors. The annotation schema defines the label space and its reliability (e.g., inter-annotator agreement), the supervision regime determines whether models operate in zero-shot or are adapted via fine-tuning, and the prompting strategy specifies how predictions are structured (direct vs. hierarchical decomposition). In particular, the annotation schema should not be conflated with hierarchical prompting: while both introduce structure, the former concerns label design, whereas the latter defines the inference procedure applied to a fixed label space. Prior work typically varies these factors independently or implicitly. By controlling supervision and holding label information constant, we isolate the effect of hierarchical decomposition and show how its benefits depend on annotation noise and model adaptation. Our contributions are primarily methodological and threefold:
\begin{enumerate}
    \item We introduce an intent-based taxonomy of propaganda techniques, the first entirely new conceptual taxonomy since 2019;
    \item We conduct a controlled comparison of our custom schema and the recent taxonomy from  \citet{sahitaj-etal-2025-hybrid} and error analysis of results. For this, we also create high level labels, using a new clustering procedure, resulting in the highest IAA in human validation for propaganda annotation;
    \item For both schemas, we compare 4 models (Phi-4, GPT-4.1-nano, Qwen2.5 and 3) and two approaches to high-level classification: \textbf{Direct-High}, where models predict high-level categories directly, and the hierarchical \textbf{Main\textrightarrow High} strategy, where fine-grained labels are predicted first and then aggregated.
\end{enumerate}
Our results and through error analysis show that fine-tuning is decisive for improving performance across models. We also demonstrate that our hierarchical propaganda prompting (\textbf{HiPP}) method improves high-level classification for our schema once models are fine-tuned, offering a stable methodological advantage in settings with noisy and imbalanced annotations. Overall, the contribution lies in analysing how different factors interact.  Our data, labels and code are available in a Github\footnote{\url{https://github.com/verosol/propaganda_hierarchical}}. Best performing fine-tuned models are accessible in a Hugging face repository\footnote{\url{https://huggingface.co/collections/xplainlp/propaganda-classification}}.


\section{Related Work}
\textbf{Datasets}. \citet{da-san-martino-etal-2019-fine} introduced the first dataset for fine-grained propaganda detection. Their schema introduced  18 propaganda techniques (e.g. loaded language, name-calling, appeal to fear) annotated at the span level, which became standard ever since, with minimal revisions (e.g. condensing the schema to 10 \cite{Kyslyi2025UNLP} or 14 labels by \citet{da-san-martino-etal-2020-semeval, sprenkamp2023large, abdullah2022detecting}). At the same time, subsequent work indicated high subjectivity and complexity of the annotation process \cite{Maarouf_2024, sahitaj-etal-2025-hybrid}.\\ 
\textbf{Propaganda classification}. Recent work has explored LLMs for propaganda detection. \citet{sahitaj-etal-2025-hybrid} proposed a hybrid annotation framework that leverages LLMs to pre-annotate texts, followed by human validation. This approach uses a hierarchical taxonomy of 14 techniques grouped into three broader categories, and demonstrates that LLM pre-annotations improve both inter-annotator agreement and annotation efficiency. 
 \citet{jose2025large} evaluated GPT-3.5, GPT-4, and Claude on six techniques in news articles, while \citet{hasanain2023large} used GPT-4 to generate multilabel and sequence tagging annotations for 23 techniques in Arabic.  \citet{sprenkamp2023large} benchmarked GPT-3 and GPT-4 for multi-label classification, showing close to SOTA performance.
 Recent work has begun to move beyond surface-level detection toward deeper analysis of propaganda \cite{liu-etal-2025-propainsight} and LLM explanations \cite{hasanain-etal-2025-propxplain}. In contrast to these approaches, our work focuses on how annotation schema design and prompting strategy interact to shape learnability and robustness in high-level propaganda classification.


\section{Methodology and Experimental setup}
\subsection{Datasets and Annotations}

We used the HQP benchmark dataset \cite{Maarouf_2024} to co-annotate tweets from \citet{sahitaj-etal-2025-hybrid}. It contains 30,000 English tweets annotated with fine-grained propaganda techniques, collected at the start of the Russian war on Ukraine. 

To capture the dynamics of modern online propaganda, we developed a new set of labels using an iterative, mixed-methods approach combining expert annotation, LLM-assisted harmonization, and group consensus. Five annotators with advanced backgrounds in computer science and familiarity with propaganda research, and independently labeled 300 tweets from the HQP dataset, describing the intent of each message in free text rather than selecting from predefined categories. 

These descriptions were clustered using GPT-4 to identify overlapping concepts and refined in group discussion, where we additionally grounded the resulted categories in \citet{Starbird2019Disinformation} and \citet{Entman1993Framing} works. This resulted in 17 fine-grained labels, which were also organized into five higher-level categories, forming a two-tier taxonomy (Table~\ref{tab:labels_low} and \ref{tab:labels_high}; Appendix~\ref{app:taxonomy} for the complete label set).
We then validated and refined the schema on a larger sample of 500 tweets, where each tweet was annotated on three levels: (i) all fine-grained techniques present, (ii) the single most prominent technique, and (iii) the overarching high-level intent. This process yielded the final hierarchical labeling schema, designed to support both nuanced analysis and robust high-level classification.

\subsection{Inter-Annotator Agreement}\label{subsec:iaa}

To gauge annotation reliability, we computed inter-annotator agreement (IAA) on the propaganda portions of the golden test sets. As shown in Table~\ref{tab:iaa-overall}, our schema yielded only moderate agreement ($\kappa{=}0.309$, $\alpha{=}0.308$), whereas \citet{sahitaj-etal-2025-hybrid}’s schema achieved substantially higher agreement ($\kappa{=}0.594$, $\alpha{=}0.594$). 

The gap highlights a fundamental trade-off. \citet{sahitaj-etal-2025-hybrid}’s categories are easier to apply consistently, yielding stronger agreement. By contrast, our intent-focused taxonomy is cognitively more demanding and introduces genuine ambiguity: annotators often disagreed not because of unclear guidelines, but because multiple propagandistic intents were simultaneously plausible. This reflects the inherent difficulty of labelling intent in short, noisy social media texts.
Across framing-style annotation campaigns, to which our scheme is arguably closer than to traditional propaganda taxonomies, reported raw inter-annotator agreement is typically low to moderate (e.g., Krippendorff’s $\alpha \approx 0.3$–$0.4$ for generic frames and $\approx 0.1$–$0.2$ for span-level unitizing; \citealp{card-etal-2015-media,piskorski-etal-2023-semeval,bassi-etal-2025-annotating}). Prior work consistently attributes this to the inherent subjectivity and multi-label nature of framing, and treats IAA primarily as a signal for managing annotators and consolidation rather than a strict quality threshold.

\begin{table}[t]
  \centering
  \begin{tabular}{lcc}
    \toprule
    \textbf{Schema} & \textbf{Cohen's $\kappa$} & \textbf{Krippendorff's $\alpha$} \\
    \midrule
    Our scheme            & 0.309 & 0.308 \\
    Sahitaj et al. & 0.594 & 0.594 \\
    \bottomrule
  \end{tabular}
  \caption{Overall inter-annotator agreement on propaganda tweets (main labels).
  }
    \label{tab:iaa-overall}
\end{table}


\subsection{Schema Comparison}
We consider two schemas, each with \textbf{17} propaganda techniques plus one non-propaganda class, grouped into \textbf{six high-level} categories.

\citet{sahitaj-etal-2025-hybrid}'s schema is
a span-based taxonomy developed for news and adapted to Twitter. For comparability at the high level, we \emph{derive} six umbrellas by clustering their 17 propaganda techniques $+$ one non-propaganda technique using a similarity matrix combining label co-occurrence (Jaccard) and Dawid--Skene \cite{dawid-skene} posterior profiles; we fix the non-propaganda label to its own group (details in App.~\ref{app:clustering}).
Four annotators (A1-A4) then re-labelled a 200-tweet gold set with these six categories while seeing the original span-level annotations. We obtain substantial strict agreement (overall overlap $0.66$, Krippendorff's $\alpha_\text{nom}=0.78$). The clustering-based labels (A5) match the human majority vote with macro-/micro-F1 of $0.93/0.91$ (Tab.~\ref{tab:cluster_iaa}), indicating that the induced high-level schema is stable and reproducible.
\begin{table}[t]
\centering
\begin{tabular}{lc}
\toprule
Metric & Score \\
\midrule
Overall overlap (A1--A4)         & 0.66 \\
Krippendorff $\alpha_\text{nom}$  & 0.78 \\
A5 vs.\ majority vote macro-F1    & 0.93 \\
A5 vs.\ majority vote micro-F1    & 0.91 \\
\bottomrule
\end{tabular}
\caption{Validation of clustered high-level schema on 100 tweets.
A1--A4: human annotators; A5: clustering-based labels.
Strict uses the primary label only; lenient counts agreement if either
primary or secondary label matches.}
\label{tab:cluster_iaa}
\end{table}
\subsection{Models}

All experiments use \textbf{(i) GPT-4.1-nano}\cite{openai2023gpt4} (release: 2025-04-14) base and fine-tuned with the OpenAI API and provider-standard optimization defaults; and \textbf{(ii) Phi-4 14B}\cite{abdin2024phi4technicalreport} models, base and fine-tuned with LoRA-adapters; (iii-iv) Qwen2.5-14B \cite{yang2024qwen25} and Qwen3-14B \cite{yang2025qwen3} base and fine-tuned. The fine-tuning details are documented in Appendix~\ref{app:ft}.
For each schema, we used a fixed split totaling \textbf{500 tweets} (all drawn from the HQP corpus): training set (210 tweets, with 105 propaganda, 105 non-propaganda); validation set (90 tweets, with 45 propaganda, 45 non-propaganda); golden test set (200 tweets, with 100 propaganda, 100 non-propaganda). 

Non-propaganda items are identical across schemas, while propaganda items differ to preserve balanced label distribution, but all tweets are drawn from the same source corpus. The golden test set is balanced at the binary level, but distributions across fine-grained propaganda techniques remain slightly skewed. 
\begin{table*}[t]
\centering
\small        
\setlength{\tabcolsep}{3pt}
\begin{tabular}{lllcccccccc}
\toprule
 &  &  & \multicolumn{2}{c}{GPT-4.1-nano} &
 \multicolumn{2}{c}{Phi-14B} &
 \multicolumn{2}{c}{Qwen2.5-14B} &
 \multicolumn{2}{c}{Qwen3-14B} \\
\cmidrule(lr){4-5}\cmidrule(lr){6-7}\cmidrule(lr){8-9}\cmidrule(lr){10-11}
Schema & Regime & Strategy
& $M$ & $W$ & $M$ & $W$ & $M$ & $W$ & $M$ & $W$ \\
\midrule
\multirow{4}{*}{Ours}
 & Base & Direct  & 0.204 & 0.379 & 0.379 & 0.554 & 0.301 & 0.536 & 0.334 & 0.483 \\
 & Base & M$\to$H & 0.137 & 0.225 & 0.281 & 0.399 & 0.259 & 0.457 & 0.189 & 0.333 \\
 & FT   & Direct  & 0.233 & 0.477 & 0.375 & \textbf{0.562} & \textbf{0.431} & \textbf{0.630} & \textbf{0.539} & \textbf{0.685} \\
 & FT   & M$\to$H & \textbf{0.376} & \textbf{0.571} & \textbf{0.412} & 0.554 & 0.416 & 0.605 & 0.431 & 0.614 \\
\midrule
\multirow{4}{*}{Sahitaj}
 & Base & Direct  & 0.271 & 0.405 & 0.421 & 0.439 & 0.479 & 0.554 & 0.424 & 0.519 \\
 & Base & M$\to$H & 0.284 & 0.429 & 0.364 & 0.195 & 0.360 & 0.192 & 0.374 & 0.373 \\
 & FT   & Direct  & 0.441 & 0.574 & 0.493 & 0.511 & 0.498 & 0.612 & 0.448 & 0.597 \\
 & FT   & M$\to$H & \textbf{0.529} & \textbf{0.623} &
                  \textbf{0.548} & \textbf{0.602} &
                  \textbf{0.560} & \textbf{0.657} &
                  \textbf{0.560} & \textbf{0.661} \\
\bottomrule
\end{tabular}
\caption{Results across configurations. Scores are macro-F1 ($M$) and weighted-F1 ($W$).
Strategies: Direct = Direct-High, M\textrightarrow H = Main\textrightarrow High.
Regimes: Base = zero-shot baseline, FT = fine-tuned.
For Sahitaj et~al.\ (\citeyear{sahitaj-etal-2025-hybrid}) the \textbf{HiPP}
Main\textrightarrow High strategy consistently gives the best results after
fine-tuning, and absolute scores are higher overall due to stronger IAA.}
\label{tab:results_all}
\end{table*}
\begin{figure*}[h]
  \centering
  \includegraphics[width=365pt]{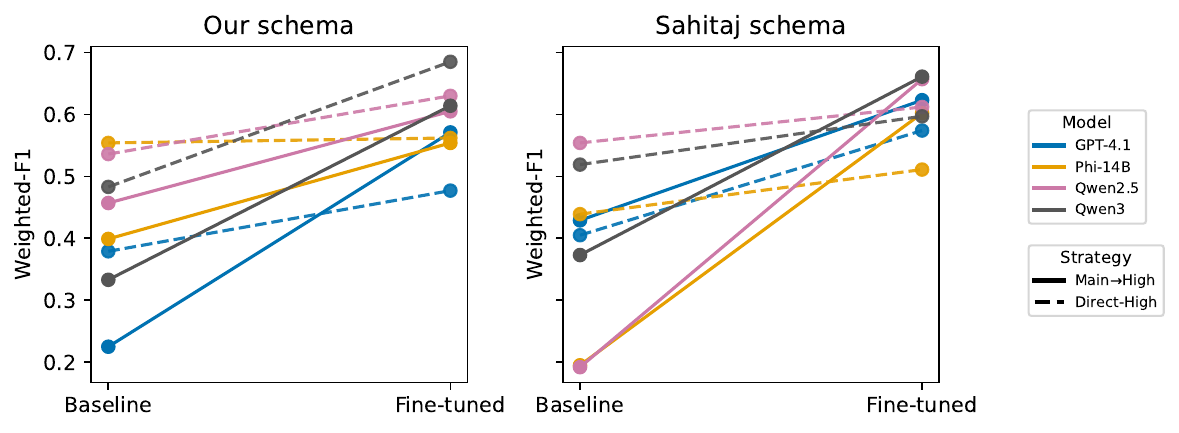}
  \vspace{-0.5em}
  \caption{\textbf{Weighted-F1 improvements from Direct-High to
  Main-High.}
  For each backbone, lines connect zero-shot (Baseline) and fine-tuned scores.
  On both schemata, Main\textrightarrow High lags behind Direct-High in zero-shot
  but gains more from fine-tuning, especially on our lower-IAA taxonomy.
  (GPT-4.1 here refers to the GPT-4.1-nano variant.)}
  \label{fig:slope_macro}
\end{figure*}

\subsection{Prompting strategies}
We compare: (i) \textbf{Direct-High} (predict a high-level category directly from text $+$ category descriptions) and (ii) our \textbf{HiPP} method, which is \textbf{Main\textrightarrow High} (first predict the prominent technique, then the high-level category in the same pass, conditioned on the main prediction).

\paragraph{Evaluation.}
We report macro-F1 and weighted-F1. Macro-F1 highlights performance on minority classes, which is critical under the imbalanced technique distributions observed in HQP; weighted-F1 reflects prevalence.

\section{Results}
\label{sec:results}

We evaluate models across both schemas and both prompting strategies. 
Table~\ref{tab:results_all} reports results for both baseline (zero-shot) and fine-tuned conditions. 
Across the board, fine-tuning substantially improves absolute performance,
while preserving the qualitative trends observed in the baselines.
Three main patterns emerge.

\paragraph{Effect of fine-tuning and prompting.}
Fine-tuning is decisive for all configurations: it yields gains of
$+0.09$--$+0.30$ weighted-F1 compared to zero-shot.
\textbf{HiPP} (Main\textrightarrow High) tends to underperform Direct-High
in the Base regime, especially on our lower-IAA schema, but benefits
disproportionately from fine-tuning.
For the Sahitaj schema, FT Main\textrightarrow High is the best configuration
for \emph{every} backbone.
On our schema, FT Main\textrightarrow High is clearly best for GPT-4.1-nano
and competitive for the other models, while the stronger configurations obtain
their top scores with FT Direct-High.
These results indicate that the effectiveness of HiPP depends strongly on the interaction between supervision regime and label-space characteristics. In the zero-shot (Base) setting, Main→High often underperforms Direct-High, suggesting that hierarchical decomposition can amplify noise when model representations are not adapted. After fine-tuning, however, Main→High consistently improves or matches Direct-High, particularly on the lower-IAA schema, indicating that hierarchical prompting becomes beneficial once the model has learned task-specific representations. Rather than depending solely on model capacity, the observed gains arise from the interaction between supervision and label reliability.
Overall, these results show that hierarchical prompting is not universally beneficial, but relies on fine-tuning to provide gains.

\paragraph{Effect of schema / IAA.}
Absolute performance is consistently higher on the Sahitaj schema across
all models and strategies, with gaps of roughly $+0.05$--$+0.15$ in weighted-F1.
This aligns with its substantially stronger inter-annotator agreement and supports our claim that
our intent-based taxonomy is the more challenging setting.
The fact that Main\textrightarrow High excels on both schemata after fine-tuning
indicates that the \textbf{HiPP} strategy provides a robust methodological
advantage and is not merely overfitting to a specific label design.
Even under strong annotator disagreement, the fine-tuned
Main\textrightarrow High model can recover the majority-vote label.
For one tweet, annotators split between Shift Blame (2 votes),
Distort Reality (2), and Delegitimise (3), yielding
Krippendorff's $\alpha = 0.18$.
Only the Main\textrightarrow High configuration predicted the
majority label correctly; the full example is given in
Appendix~\ref{app:unique-example}.

\paragraph{Model comparison.}
Among the open-weight models, the two Qwen models achieve the highest
scores overall, with Qwen3-14B reaching 0.685 $W$-F1 on our schema and
0.661 on Sahitaj.
Phi-4 14B consistently outperforms GPT-4.1-nano in both regimes and on
both schemata, confirming that our findings are not specific to a single
model family but hold across multiple moderately sized LMs. Weighted-F1 scores are systematically higher than macro-F1, reflecting
the skewed label distribution: frequent classes dominate weighted-F1,
while macro-F1 highlights the remaining difficulty in rarer intents.

\section{Error Analysis}

\subsection{Setup}
We perform a qualitative error analysis that combines confusion matrices with lemma-frequency statistics. For the latter, we lemmatise all test tweets using the spaCy large English \footnote{\texttt{en\_core\_web\_lg} (\url{https://spacy.io/models/en})}, then group them into true positives, true negatives, false positives, and false negatives, separately for each model, training regime (Base vs.\ FT), and schema (ours vs.\ \citet{sahitaj-etal-2025-hybrid}). For every confusion-cell we compute the most frequent lemmatised tokens. Full confusion matrices and lemma tables are provided in Github repository and example for Qwen3 results is given in Appendices~\ref{app:token-files-example} and~\ref{app:qwen3-confusions}. 

\begin{figure*}[t]
  \centering
  \begin{subfigure}[b]{0.45\textwidth}
    \centering
    \includegraphics[width=\linewidth]{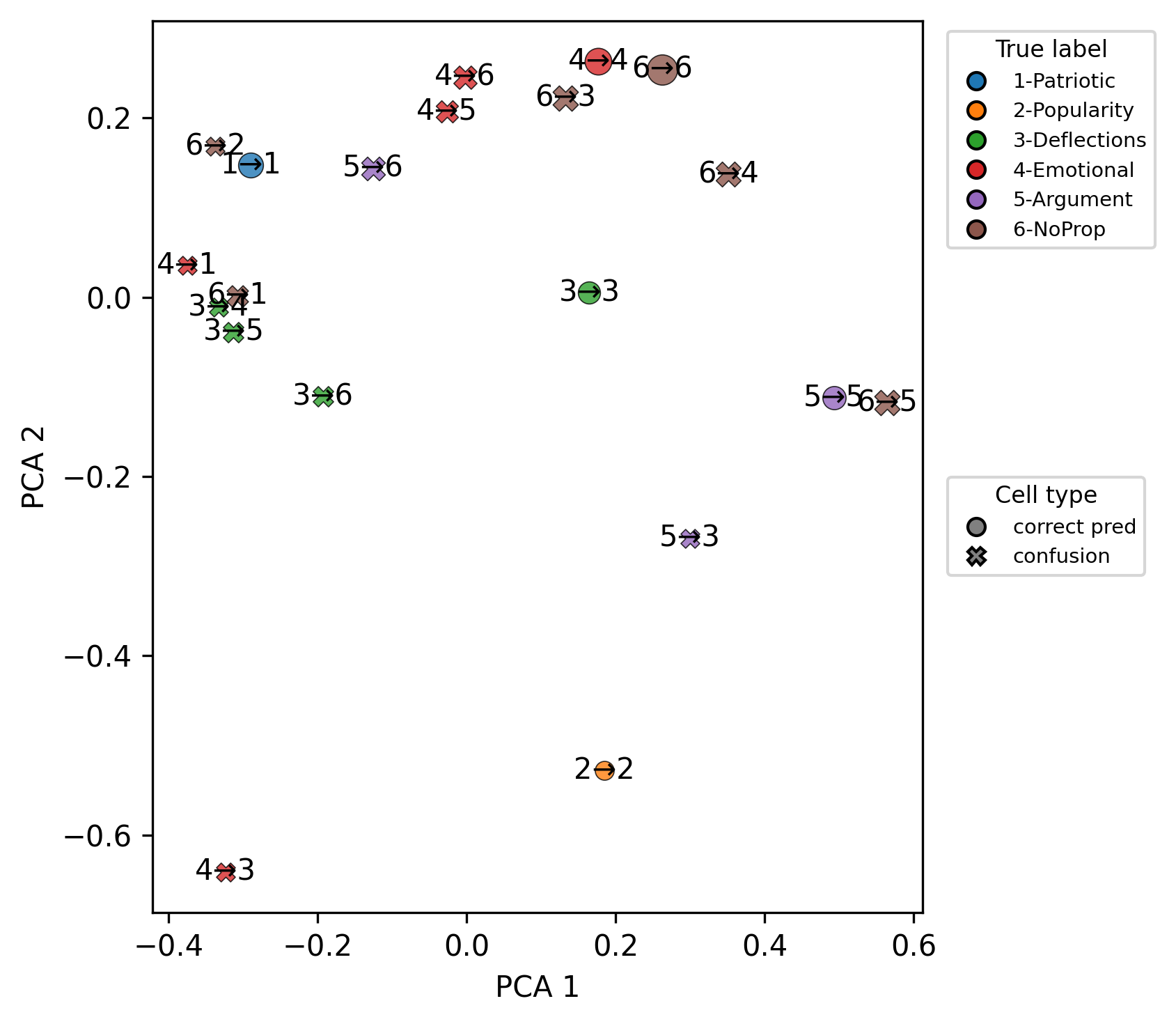}
    \caption{Sahitaj high-level clusters}
    \label{fig:pca_sahitaj}
  \end{subfigure}
  \hfill
  \begin{subfigure}[b]{0.45\textwidth}
    \centering
    \includegraphics[width=\linewidth]{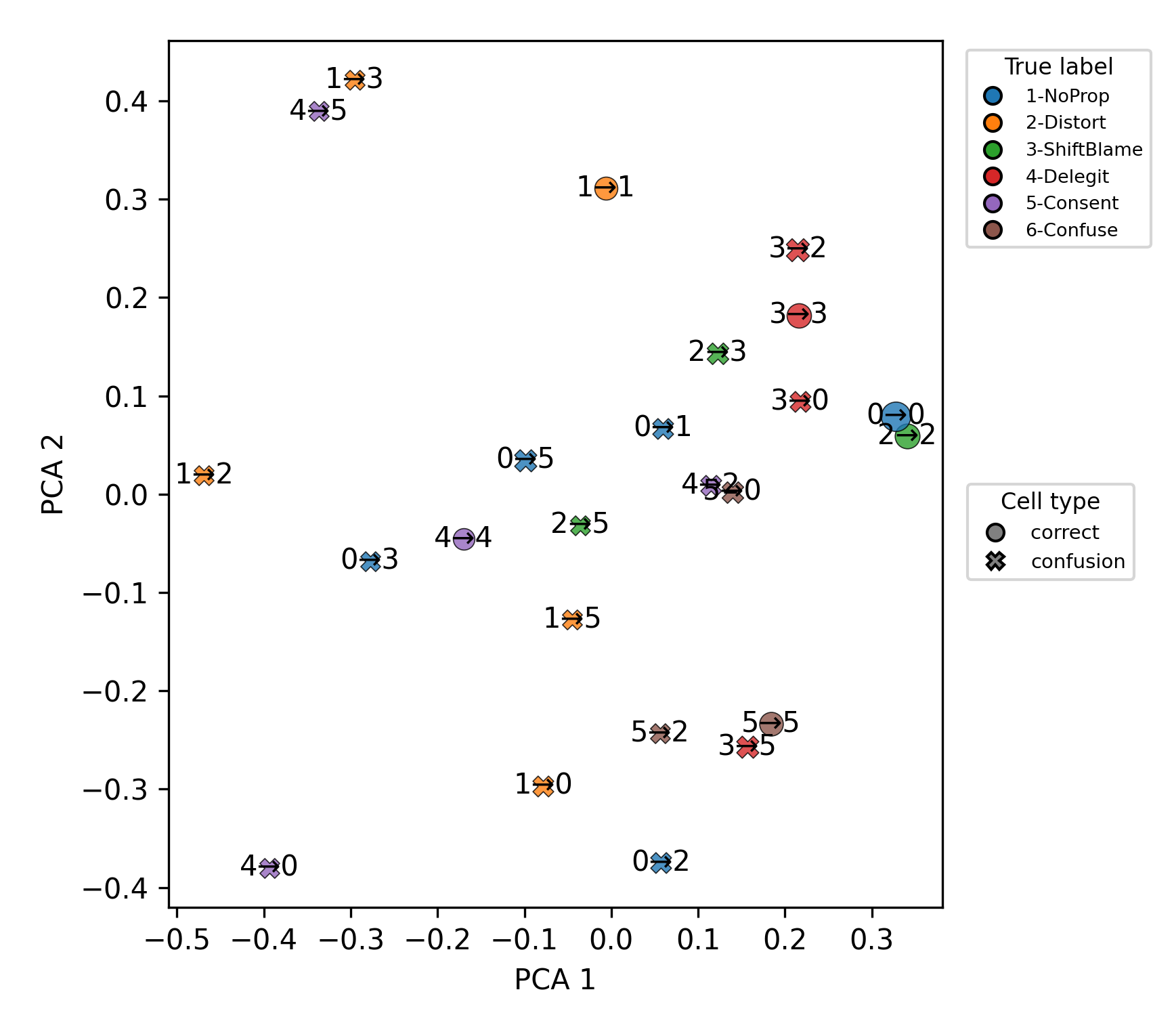}
    \caption{Our intent-based high-level schema}
    \label{fig:pca_ours}
  \end{subfigure}
  \caption{PCA of confusion-cell TF--IDF vectors for Qwen3-14B in its best setting. 
  Each point represents a (true, predicted) high-level label cell; colours indicate 
  true labels and markers distinguish correct vs.\ confused cells.}
  \label{fig:pca_both}
\end{figure*}

We also analyse our best results for Sahitaj and our scheme (both Qwen3-14B), by visualising all confusion cells (true, predicted) as TF–IDF vectors and projecting them to two dimensions with PCA (Figs.~\ref{fig:pca_both}; we include all cells, even those with only a single instance). Since we are interested in the global variance within the datapoints, we apply PCA instead of other techniques like t-SNA or UMAP.

Our analysis focuses on two research questions: (i) non-propaganda vs.\ propaganda confusions, which dominate the error mass, and (ii) within-propaganda confusions between high-level intent and technique categories.
\subsection{Non-Propaganda vs. Propaganda}
Across all models, schemas and regimes, correctly and incorrectly classified non-propaganda tweets share the same topical lexicon: ukraine, russia, war, people, crisis are among the most frequent lemmas in true non-propaganda (0\textrightarrow  0). At the same time, almost identical vocabularies appear in non-propaganda tweets that are misclassified as propaganda. The key difference is strength of framing. Mislabelled instances contain additional high-intensity cues such as nazi/nazis, genocide, terrorist, destroy, ban, closethesky, sendnatotoukraine, or insults and culture-war terms (e.g., fuck, woke).

In the Base Direct-High regime, this leads all four models to be hyper-sensitive to these keywords: any combination of ukraine/russia/war with Nazi/WWII rhetoric is frequently mapped to attack/shift-blame/emotional category, even when the tweet is reporting or critical commentary. In the Base Main\textrightarrow  High setting, this effect is amplified: once a main technique is chosen, war+Nazi lexicon tends to be “locked in” as a propaganda high-level label, inflating 0\textrightarrow  propaganda errors.

Fine-tuning, and especially fine-tuned Main\textrightarrow  High, sharply reduces this behaviour. For both schemas, the first row of the confusion matrices becomes much more diagonal: neutral war reporting, often including URLs and event descriptors, is kept in class 0, while the remaining 0\textrightarrow  propaganda mass concentrates on borderline, highly polarised cases such as explicit genocide claims or accusations of neo-Nazi identity. This pattern holds for GPT-4.1-nano, Phi-4, Qwen2.5 and Qwen3.

\subsection{Confusion between propaganda classes}
\subsubsection{Our Schema}
For our intent and framing based schema, propaganda labels (1–5) are mainly confused with adjacent goals in the hierarchy.
Distort-Reality tweets misclassified as Shift Blame (1\textrightarrow  2) typically mix historical reinterpretation (coup, election, proxy war, Donbass) with responsibility-shifting (NATO, Biden, West), making the boundary between revising the past and justifying current aggression inherently fuzzy. Shift-Blame tweets mislabelled as Delegitimise (2\textrightarrow  3) are dominated by dehumanising rhetoric (nazi, battalion, hitler, swastika, salute), so models focus on moral condemnation rather than the blame-transfer structure.
 Manufacture-Consent posts mispredicted as Shift or Delegitimise (4\textrightarrow  2/3) combine in-group mobilisation (istandwithputin, istandwithrussia, bandera, ukronazis) with explicit attacks on the out-group. Models sometimes prioritise the attack aspect over the identity-building goal.
Confuse \& Distract (5) exhibits the most heterogeneous lexicon (paypal, ban, twitter, donation, tank, weapon, western media), and its instances leak into several other propaganda categories. This suggests that cues for distraction and fragmentation (multiple topics, logistics, meta-commentary) are not well captured by simple lexical patterns.

Fine-tuning makes these confusions more taxonomically local. In FT Direct-High, correctly classified Shift-Blame items concentrate on expected frames (nazi, nato, victimhood), while misclassified ones tend to mix elections, coups and moral condemnation. In FT Main\textrightarrow  High, off-diagonal mass is largely restricted to Distort$\leftrightarrow$Shift and Shift$\leftrightarrow$Delegitimise; long-range errors (e.g., Distort$\leftrightarrow$Confuse) are strongly reduced.

\subsubsection{Sahitaj scheme}
For the clustered Sahitaj schema, the high-level groups already merge multiple techniques, and lemma-based errors reflect multiple tendencies. Non-propaganda is often mislabelled as Emotional \& Loaded Persuasion or Deflections \& Distractions. This is driven lexicon such a \textit{nazi/nazis, threat, destroy, bomb, ban, closethesky} and other standard war lexicon. Within propaganda classes, confusions concentrate on (i) Emotional $\leftrightarrow$ Deflections, when strong affect co-occurs with causal or whataboutist reasoning (\textit{nazi, propaganda, fake, fear, attention}) and (ii) Deflections $\leftrightarrow$ Argument Manipulations, when reframing causality is combined with thought-terminating clichés or straw-man attacks.

Compared to our schema, the clustered Sahitaj categories show cleaner lexical separations. Popularity and patriotic appeals are tightly tied to slogan hashtags (istandwithrussia, standwithputin, sendnatotoukraine), while Argument Manipulations collect more abstract reasoning tokens (logic, lie, believe, question). After fine-tuning, Qwen2.5 and Qwen3 in particular handle the patriotic class almost perfectly, with very few leaks into other clusters.

\subsection{Model and Regime Effects} 

 Phi-4’s errors are lexically the sharpest. When it misclassifies non-propaganda as Deflections, the top tokens overwhelmingly support that reading (threat, propaganda, nato, stop, biden). GPT-4.1-nano is the noisiest, spreading the same war vocabulary across more categories. Qwen2.5 and Qwen3 sit in between: their error lexicons are more concentrated than GPT-4.1-nano’s, but still less focused than Phi-4’s. Qwen3 is slightly more conservative post-FT, with especially clean non-propaganda rows and crisp handling of slogan-driven patriotic appeals.
 Base models are much more sensitive to simple lexical cues such as “war" and "Nazi”, regardless of context. Fine-tuning shifts errors toward genuinely ambiguous, polarised tweets and shrinks lemma distributions in error cells: generic war terms recede, while extreme cues (\textit{nazi, genocide, ban, closethesky}) dominate the remaining mistakes.
 
 In zero-shot, Main$\,\to\,$High increases false positives. Once a main technique is predicted, the model tends to “commit” to a propaganda label whenever strong lexical cues are present. After fine-tuning, the same hierarchical structure becomes stabilising, since the results show that non-propaganda is rarely assigned any technique at the MAIN level, and HIGH-level errors align with conceptual neighbourhoods (Distort-Shift, Emotional $\leftrightarrow$ Deflections $\leftrightarrow$ Argument).

 \subsection{Performance of clustering-based labels}
The clustered high-level labels derived from \citet{sahitaj-etal-2025-hybrid} turn out to be both annotation- and model-friendly. On the human side, they achieve substantial strict agreement and very high lenient agreement in our human validation, indicating that annotators can reliably reason when given the underlying span-level techniques. On the modelling side, several clusters behave strikingly well across all models and regimes. Patriotic \& Catchy Appeals is almost a “slogan class”: after fine-tuning, tweets with dense campaign-style hashtags (e.g., \textit{istandwithputin}, \textit{istandwithrussia}) are classified with very high precision and recall. No Propaganda is the main beneficiary of fine-tuning with \textbf{HiPP}: while zero-shot models often over-label war content as propagandistic, FT Main\textrightarrow High yields a strong diagonal for this class, keeping neutral, URL-rich reporting in NP and relegating only the most polarised war+Nazi posts to propaganda. Within propaganda, Emotional \& Loaded Persuasion, Deflections \& Distractions, and Argument Manipulations form a tight triad: they achieve good F1 scores, and their remaining errors are largely confined to confusions inside this trio (e.g., descriptive Emotional content drifting into Deflections, or reasoning-heavy Deflections drifting into Argument Manipulations). Taken together with the high co-annotator agreement, this suggests that the clustered Sahitaj schema is coarse enough to be robust and learnable, yet structured enough to preserve meaningful distinctions in propagandistic style.
\subsection{PCA analysis}
In both schemata, cells sharing the same true label form compact lexical neighbourhoods, and confusions stay close to the corresponding correct cell, indicating that errors are locally coherent rather than arbitrary. For the clustered Sahitaj labels, these neighbourhoods are tight and well separated, whereas for our schema the Distort, ShiftBlame, Delegitimise and Consent intents occupy a broad overlapping region with many small cross-label cells. In plot B, we can observe that NoProp (0) cells lie adjacent to correctly predicted ShiftBlame and Delegitimise cells (2→2, 3→3), and its false positives (0→1/2/3/5) occupy the same region, illustrating how borderline non-propaganda tweets share vocabulary with attack and responsibility-shifting propaganda. The propaganda intents Distort, ShiftBlame, Delegitimise and Consent (1–4) form a broad overlapping cluster. Confuse \& Distract (5) has no clean cluster of its own, as its cells are interspersed with ShiftBlame and Delegitimise, reflecting the heterogeneous, catch-all nature of this intent. This denser view confirms our qualitative finding that the Sahitaj clusters are easier and lexically cleaner, while our taxonomy induces a more tangled, higher-variance decision space.
\section{Discussion}

Our study examined three dimensions of propaganda detection: model portfolio selection, schema-level comparison, and prompting strategy. Across all experiments, a consistent pattern emerges: fine-tuning is decisive, and the \textbf{HiPP} Main\textrightarrow High strategy is particularly beneficial once models have been adapted to the task.

Zero-shot performance varies widely across models and strategies, whereas fine-tuning reliably turns weak baselines into competitive systems and makes methodological effects visible that are almost invisible in the Base regime. This suggests that base models substantially underestimate the potential of open-weight LMs for specialised tasks such as propaganda detection.

Schema design and reliability also shape outcomes. Models trained on \citet{sahitaj-etal-2025-hybrid}’s higher-agreement schema ($\kappa{=}0.594$) consistently outperform those trained on our intent-focused, lower-agreement taxonomy ($\kappa{=}0.309$). This illustrates a fundamental trade-off: simplified, technique-centric schemata provide clearer learning signals, while more ecologically realistic, intent-based schemata better reflect real-world complexity but introduce ambiguity and label noise. However, the best settings only differ in 0.02 points making both schema application possible, depending on the application use case, without strong performance trade-off.

Among the open-weight models we test, Phi-4 14B already outperforms GPT-4.1-nano across schemata, and the two Qwen models achieve the best absolute scores, with Qwen3-14B reaching 0.685 weighted-F1 on our schema and 0.661 on Sahitaj. 
The benefits of HiPP are most pronounced after fine-tuning and in the presence of noisier label spaces, rather than being determined solely by backbone strength: for GPT-4.1-nano, fine-tuned Main\textrightarrow High outperforms Direct-High on both schemata, improving weighted-F1 by +0.094 (ours) and +0.049 (Sahitaj). For stronger models, Main\textrightarrow High remains the best configuration on the Sahitaj schema and is competitive on ours, while some models obtain slightly higher scores with Direct-High. Overall, this pattern suggests that \textbf{HiPP} provides a useful inductive bias for coping with noisy, imbalanced annotations, especially when the label space is challenging or model capacity is limited, without requiring changes to the underlying schema. However, as both Direct-High and Main→High observe the same supervision signal, performance gains can be attributed to hierarchical decomposition rather than increased label information.

Overall, the analysis suggests that all four models, across both taxonomies, implement a similar inductive bias: they strongly over-weight topic + affective markers (Ukraine/Russia/war + Nazi/WWII rhetoric) as signals of propaganda, and they treat high-level propaganda intents in a locally coherent way, mostly confusing neighbouring goals rather than arbitrarily jumping across the taxonomy. Fine-tuning in the \textbf{HiPP} Main\textrightarrow  High regime turns these models from crude lexical detectors into structured classifiers whose residual mistakes are concentrated in the most contentious cases—precisely where human annotators also tend to disagree.

At the same time, our intent-based schema remains visibly harder than the clustered Sahitaj schema. Even after fine-tuning, several of our propaganda classes remain mutually confused, mirroring the lower human IAA we observe. From a practical perspective, this highlights a trade-off between expressive taxonomies and clustered high-level schema, which is easier to learn and deploy but less informative, making it less suitable for application on real-world data. Our results indicate that, when fine-tuned with HiPP, modern LLM models from different families converge to very similar and interpretable error structures, providing some reassurance that conclusions drawn from one model reasonably transfer to others.

The PCA visualisations of confusion cells from the best settings also highlight how our taxonomy reshapes the decision space. While the Sahitaj clusters yield compact, well-separated lexical neighbourhoods with mostly local errors, our intent-based schema produces a much denser, overlapping landscape, suggesting that the added expressivity of our labels comes at the cost of a more difficult and inherently noisier classification problem. In this context, we believe that our proposed taxonomy has substantial value. Beyond methodological findings, our intent-focused taxonomy opens new opportunities for research and practice. By emphasizing communicative goals rather than surface-level techniques, it enables nuanced studies of how propaganda strategies evolve across conflicts and cultures, and how they interact with public opinion and polarization. For practitioners, it can support early-warning systems that focus not just on whether a message is propaganda, but also why it was crafted and how it seeks to influence audiences.
At the same time, high level labels we created for Sahitaj et al. scheme, show the most promise in terms of IAA. Although validated on a limited sample, they show one of the highest agreement results seen so far in literature.

\section{Conclusion}
In this work, we introduced a new hierarchical labeling schema for online propaganda, focused on intent and key message, and conducted a controlled comparison between our taxonomy and the recent framework by \citet{sahitaj-etal-2025-hybrid}. We evaluated how schema design, annotation reliability, and prompting strategies interact, comparing various similar sized models, direct high-level classification and a \textbf{HiPP} approach. Our results show that while schema reliability shapes performance, LLM fine-tuning and \textbf{HiPP} provides an advantage, offering robust supervision specifically when annotation quality is imperfect or datasets are heterogeneous.
\section*{Limitations}
Our study has several limitations. First, it was conducted on relatively small subsets of the HQP corpus (500 items per schema), with results based on a single split. Testing across two schemas and two models helped mitigate setup-specific variance, but future work should confirm stability under multiple random splits. Given the relatively small dataset size and single train/test split, the reported results should be interpreted as indicative of methodological trends rather than definitive performance estimates. 
Second, our clustering of Sahitaj et al.'s \cite{sahitaj-etal-2025-hybrid} main labels into high-level groups involved interpretive choices (e.g., $\lambda$ for similarity weighting).  Future work should therefore scale up evaluation, test cross-validation setups, and replicate high-level clustering with alternative methods or expert-driven designs. 
It will also be valuable to explore hierarchical prompting in multi-task or multi-step architectures, and to design hybrid corpora that balance ecological realism with annotation reliability. 

Exploring additional open-weight and proprietary models and their larger and smaller versions would help better generalize the trends we found in this study.


Finally, our labeling schema operates at the intersection of \textit{framing} and \textit{propaganda}. Future work will focus on disentangling these two phenomena within one schema to better estimate which part is more challenging for LLMs.

Our annotators were primarily early-career researchers with backgrounds in computer science from Germany and Ukraine; while this provided valuable regional expertise, broader demographic coverage would further strengthen future work.

\section*{Ethical Considerations}
The annotation process inherently reflects subjective judgments, and schema design choices may encode cultural or political biases. To mitigate this, we used a multilingual team of seven annotators and conducted consensus discussions. Nevertheless, our models may still exhibit bias toward particular geopolitical or linguistic contexts, and results should be interpreted with caution.

Finally, while our methods are intended to support fact-checking and disinformation monitoring, they could be misused for large-scale surveillance or censorship. We emphasize that detection tools should always be deployed in combination with human oversight and transparent governance structures.

While working on the manuscript, we have used LLM assistants for the purpose of spell-checking and as a writing assistant, but not for the creation of the content.   

\section*{Acknowledgments} 
The work on this paper was performed in the scope of the projects “VeraXtract” (16IS24066) and “news-polygraph” (reference: 03RU2U151C) funded by the German Federal Ministry for Research, Technology and Aeronautics (BMFTR).

\bibliography{custom}

\appendix

\section{Representative Low-IAA Example}
\label{app:unique-example}

\noindent\textbf{Tweet (full).}
\begin{quote}\itshape
@USER @USER Oh no ….another “Jews for Nazis”. Is it because that the Nazis are killing Russians and not Jews — makes it OK?
\end{quote}

\noindent\textbf{Human votes (7 raters).}
2× \textit{Distort reality and rewrite the past} (1), 
2× \textit{Shift blame and Justify Aggression} (2), 
3× \textit{Delegitimize the opponent} (3). 
The plurality gold label is therefore \textit{Delegitimize the opponent} (3/7).

\medskip

\noindent\textbf{Model predictions.}
Table~\ref{tab:lowiaa-case} compares system outputs against the gold label. 
Only fine-tuned Main\textrightarrow High predicts the majority-vote label, 
while all other strategies misclassify the tweet.

\begin{table}[th]
\centering
\small
\setlength{\tabcolsep}{6pt} 
\renewcommand{\arraystretch}{1.15} 
\begin{tabular}{l c l c}
\toprule
\textbf{Model} & \textbf{Pred (int)} & \textbf{Predicted label}  \\
\midrule
Direct (Base)   & 2 & Shift blame \& \\
                &   & Justify Aggression \\
Direct (FT)     & 2 & Shift blame \&   \\
                &   & Justify Aggression \\
M$\to$H (Base)  & 1 & Distort reality \&    \\
                &   & Rewrite past \\
\textbf{M$\to$H (FT)} & \textbf{3} & \textbf{Delegitimize opponent} \\
\midrule
\textbf{Gold (plurality)} & \textbf{3} & \textbf{Delegitimize opponent}  \\
\bottomrule
\end{tabular}
\caption{\textbf{Representative low-IAA case.} 
Annotator votes: 1. Distort reality (2/7 votes), 2. Shift blame (2/7 votes), 3. Delegitimize opponent (3/7 votes). 
Plurality gold = \textit{Delegitimize opponent} (3/7). 
Only fine-tuned M\textrightarrow H predicts the correct label.}
\label{tab:lowiaa-case}
\end{table}

\section{Clustering of Sahitaj et al.'s Techniques}\label{app:clustering}
To enable high-level comparisons, we clustered the 17 main propaganda techniques from Sahitaj et al.~\cite{sahitaj-etal-2025-hybrid} into five umbrella propaganda categories plus the non-propaganda class. 

\paragraph{Method.}
We construct a similarity matrix 
$S_{ij}=\lambda\,J_{ij}+(1-\lambda)\,DS_{ij}$, 
where $J_{ij}$ is Jaccard co-occurrence across tweets and $DS_{ij}$ is the similarity of Dawid–Skene posteriors. 
The non-propaganda label is fixed to its own group. 
To avoid data leakage, clustering was performed only on the training set.

\paragraph{Agreement Across $\lambda$.}
Table~\ref{tab:lambda-sweep} shows agreement scores across similarity weightings. 
Agreement improved as $\lambda$ approached pure co-occurrence ($\lambda=1.0$), peaking at Fleiss' $\kappa = 0.738$ and Krippendorff's $\alpha = 0.738$. 
We selected $\lambda=0.5$ as a balanced setting: clusters remained interpretable and well-sized, with only a small drop in agreement.

\begin{table}[!ht]
\centering
\small
\setlength{\tabcolsep}{2pt}  
\renewcommand{\arraystretch}{0.95} 
\begin{tabular}{c c c}
\toprule
$\lambda$ & Fleiss' $\kappa$ & Krippendorff's $\alpha$ \\
\midrule
0.0 & 0.683 & 0.684 \\
0.1 & 0.714 & 0.714 \\
0.2 & 0.719 & 0.719 \\
0.3 & 0.719 & 0.719 \\
0.4 & 0.719 & 0.719 \\
0.5 & 0.719 & 0.719 \\
0.6 & 0.719 & 0.719 \\
0.7 & 0.725 & 0.725 \\
0.8 & 0.725 & 0.725 \\
0.9 & 0.733 & 0.733 \\
1.0 & 0.738 & 0.738 \\
\bottomrule
\end{tabular}
\caption{Agreement vs.\ $\lambda$ for hybrid similarity weighting on Sahitaj et al.'s schema.}
\label{tab:lambda-sweep}
\end{table}

\paragraph{Cluster Compositions.}
The $\lambda=0.5$ solution yielded six interpretable umbrella categories. These roughly align with communicative functions such as patriotic appeals, popularity appeals, deflections, emotional persuasion, logical/argument manipulations, and non-propaganda.

\begin{itemize}
    \item Group 1: \{15, 5\}
    \item Group 2: \{16, 9\}
    \item Group 3: \{14, 17, 7, 8\}
    \item Group 4: \{12, 13, 3\}
    \item Group 5: \{1, 10, 11, 2, 4, 6\}
    \item Group 6: \{18 (non-propaganda)\}
\end{itemize}

\section{Fine-tuning Details}
\label{app:ft}

We document our fine-tuning strategies in Table \ref{tab:ft_oi} for the GPT-4.1-nano model (fine-tuned via OpenAI API) and in Table \ref{tab:ft_phi4} for the Phi4-14B model (fine-tuned via the unsloth framework \footnote{\url{https://github.com/unslothai/unsloth}}). All results are based on a single fixed split (no cross-validation). Fine-tuning ran to completion without early stopping. 

\begin{table}[!ht]
\centering
\small
\setlength{\tabcolsep}{2pt}
\renewcommand{\arraystretch}{0.95}
\begin{tabular}{@{}p{0.40\columnwidth} p{0.50\columnwidth}@{}}
\toprule
\textbf{Fine-tuning config}  \\
\midrule
Model (base) & GPT-4.1-nano (OpenAI) (release: 2025-04-14) \\
epochs & 3 \\
batch\_size & 1 \\
learning rate multiplier & 0.1 \\
\bottomrule
\end{tabular}
\caption{Details for fine-tuning GPT-4.1-nano}
\label{tab:ft_oi}
\end{table}

\begin{table}[!ht]
\centering
\small
\setlength{\tabcolsep}{2pt}  
\renewcommand{\arraystretch}{0.95} 
\begin{tabular}{@{}p{0.55\columnwidth} p{0.45\columnwidth}@{}}
\toprule
\multicolumn{2}{l}{\textbf{Fine-tuning config}} \\
\midrule
\texttt{batch\_size} & \texttt{4} \\
\texttt{gradient\_accumulation\_steps} & \texttt{4} \\
\texttt{warmup\_steps} & \texttt{5} \\
\texttt{num\_train\_epochs} & \texttt{5} \\
\texttt{learning\_rate} & \texttt{2e-4} \\
\texttt{optim} & \texttt{"adamw\_8bit"} \\
\texttt{weight\_decay} & \texttt{0.01} \\
\texttt{lr\_scheduler\_type} & \texttt{"linear"} \\
\texttt{seed} & \texttt{3407} \\
\midrule
\multicolumn{2}{l}{\textbf{LoRA / PEFT Adapter}} \\
\midrule
\texttt{r} & \texttt{16} \\
\texttt{target\_modules} & \texttt{\{q\_proj, k\_proj, v\_proj, o\_proj, gate\_proj, up\_proj, down\_proj\}} \\
\texttt{lora\_alpha} & \texttt{16} \\
\texttt{lora\_dropout} & \texttt{0} \\
\texttt{bias} & \texttt{"none"} \\
\texttt{use\_gradient\_checkpointing} & \texttt{"unsloth"} \\
\texttt{random\_state} & \texttt{3407} \\
\texttt{use\_rslora} & \texttt{False} \\
\texttt{loftq\_config} & \texttt{None} \\
\midrule
\multicolumn{2}{l}{\textbf{Model load / unsloth configuration}} \\
\midrule\texttt{from\_pretrained(model\_name)} & \texttt{"unsloth/Phi-4"} \\
\texttt{full\_finetuning} & \texttt{False} \\
\texttt{load\_in\_4bit} & \texttt{False} \\
\texttt{Final Quantization} & \texttt{8 bit} \\

\bottomrule
\end{tabular}
\caption{Hyperparameters for fine-tuning Phi4-14B, including LoRA adapter configuration}
\label{tab:ft_phi4}
\end{table}

We fine-tuned Phi-4 on 1 GPU Nvidia A6000 48Gb for approx. 2 hours each experiment, while gpt-4.1nano was fine-tuned on the OpenAI servers. The whole set of gpt-4.1nano experiments cost 40 US dollars. 

\section{New Labelling schema}
\label{app:taxonomy}

For completeness, we provide the full set of labels defined in our 
intent-focused hierarchical taxonomy. The schema consists of 
17 fine-grained propaganda techniques plus one non-propaganda class, 
which are grouped into five high-level propaganda categories and one 
non-propaganda category. Table~\ref{tab:labels_low} and Tables~\ref{tab:labels_high} restates the label inventory introduced in the main text.

Compared to prior taxonomies such as Da San Martino et al. (2019, 2020) and the hybrid hierarchical schema of Sahitaj et al. (2025), our intent-based framework shifts focus from surface-level rhetorical techniques to the underlying communicative goals of propaganda. Earlier taxonomies primarily captured stylistic and linguistic devices (e.g., loaded language, name-calling, appeal to fear), which describe how persuasion is achieved but not why a message was crafted. By contrast, our schema introduces high-level intent categories such as Shift Blame and Justify Aggression, Manufacture Consent and Identity, and Confuse and Distract, which unify multiple lower-level techniques under shared strategic purposes. This design captures manipulative intents, such as reframing moral responsibility, generating uncertainty, or mobilising identity-based polarisation, that older schemas either scattered across labels or missed entirely. As a result, the new taxonomy provides a more explanatory and goal-oriented lens for studying online propaganda, facilitating cross-context comparisons and supporting interpretability in LLM-based detection systems.
\begin{table}[ht]
\centering
\scriptsize
\setlength{\tabcolsep}{4pt}
\renewcommand{\arraystretch}{1.05}
\begin{tabular}{p{0.95\linewidth}}
\toprule
\textbf{High-Level Categories} \\ 
\midrule

\textbf{1. Distort Reality and Rewrite the Past} \newline
Goal: Undermine truth and legitimize the present through distortion. \newline
Subtypes: 2, 11, 15, 17 (False Origin Attribution, Circular Reasoning, Appeal to Ignorance). \newline
Rationale: Denial, fabrication, and invented evidence challenge fact-based narratives. \\ \midrule

\textbf{2. Shift Blame and Justify Aggression} \newline
Goal: Reframe aggressor as victim or rational actor. \newline
Subtypes: 4, 5, 13, 17 (False Causality, False Balance). \newline
Rationale: Inverts moral frameworks to rationalize wrongdoing or redirect guilt. \\ \midrule

\textbf{3. Delegitimize the Opponent} \newline
Goal: Undermine credibility and morality of adversaries. \newline
Subtypes: 1, 3, 8, 9, 12, 6. \newline
Rationale: Frames enemies as liars, extremists, or inhuman entities. \\ \midrule

\textbf{4. Manufacture Consent and Identity} \newline
Goal: Rally support and polarize identities. \newline
Subtypes: 12, 14, 16, 10, 7. \newline
Rationale: Builds loyalty through fear, pride, and tribal solidarity. \\ \midrule

\textbf{5. Confuse and Distract} \newline
Goal: Overwhelm critical thinking through noise and uncertainty. \newline
Subtypes: 7, 15, 6, 17 (Red Herrings, Appeal to Probability, False Dichotomy). \newline
Rationale: Undermines clarity, trust, and consensus. \\

\bottomrule
\end{tabular}
\caption{Overview of the five high-level propaganda categories and their purposes.}
\label{tab:labels_high}
\end{table}
\begin{table}[ht]
\centering
\scriptsize
\setlength{\tabcolsep}{4pt}
\renewcommand{\arraystretch}{1.05}
\begin{tabular}{p{0.95\linewidth}}
\toprule
\textbf{Low-Level Fine-Grained Labels} \\ 
\midrule
\textbf{1. Guilt-by-association fallacy} \newline
Most frequently as nazi analogies. Includes WWII analogies, Hitler/SS references, burning people like Nazis. Could also include other reductio ad Hitlerum instances. \\ \midrule

\textbf{2. Historical Distortion / Revisionism} \newline
Cherry-picking historical events, WWII nostalgia, Soviet glorification, fake quotes. \\ \midrule

\textbf{3. Dehumanization / Demonization} \newline
Calling opponents subhuman, monsters, or animals; atrocity narratives. \\ \midrule

\textbf{4. Deflect and Justify I: Victimhood / Gaslighting} \newline
Aggressor framed as victim, blame-shifting, moral inversion, denial of wrongdoing. \\ \midrule

\textbf{5. Deflect and Justify II: Whataboutism} \newline
False moral equivalence, hypocrisy framing (“but the West invaded Iraq”). \\ \midrule

\textbf{6. Accusation of Propaganda / Media Distrust} \newline
Asserting “the other side lies,” mocking mainstream narratives, distrust in institutions. \\ \midrule

\textbf{7. Conspiracy Narratives} \newline
CIA/Nuland plots, puppet governments, Western coups, assertion-as-proof. \\ \midrule

\textbf{8. Guilt by Association} \newline
Azov = entire military is Nazi; guilt through affiliation or past ties. \\ \midrule

\textbf{9. Sarcasm / Ridicule / Strawman} \newline
Mocking tone, exaggeration, caricatures, eye-roll emojis. \\ \midrule

\textbf{10. Emotional Manipulation / Shock Appeal} \newline
Rape/torture narratives, fearmongering, nuclear threats, mass graves. \\ \midrule

\textbf{11. False Authority / Fabrication} \newline
Fake quotes, fabricated stats, unverified claims framed as facts. \\ \midrule

\textbf{12. Us vs. Them Framing / Identity Dichotomy} \newline
Nationalist binaries, tribal solidarity, “with us or against us” rhetoric. \\ \midrule

\textbf{13. Realpolitik Framing / Moral Detachment} \newline
Cool, analytical justification of aggression; geopolitical realism. \\ \midrule

\textbf{14. Triumphalism / Victory Framing} \newline
“Russia is winning,” inevitability rhetoric, “truth will prevail.” \\ \midrule

\textbf{15. Information Laundering / Rumor Seeding} \newline
“Unconfirmed reports,” plausible deniability, sowing doubt. \\ \midrule

\textbf{16. Anti-Establishment / Anti-Globalist Framing} \newline
Criticism of NATO, EU, UN, WEF; anti-West framing. \\ \midrule

\textbf{17. Logical Fallacies} \newline
False dichotomy, causal oversimplification, circular reasoning, false balance, appeal to ignorance. \\

\bottomrule
\end{tabular}
\caption{Overview of the 17 fine-grained propaganda techniques.}
\label{tab:labels_low}
\end{table}

\clearpage
\onecolumn
\section{Token-level evidence for confusion patterns (example file)}
\label{app:token-files-example}

\begin{table*}[ht]
\centering
\small
\setlength{\tabcolsep}{6pt}
\begin{tabularx}{\textwidth}{@{} l l r X @{}}
\toprule
\textbf{True label} & \textbf{Pred label} & \textbf{\#tokens} & \textbf{Most frequent tokens (count)} \\
\midrule
Non-Propaganda (0) & Non-Propaganda (0) & 905 &
ukraine (27), war (20), russia (18), putin (9), propaganda (8), russian (8), people (7), crisis (6), live (6), uk (6) \\

Non-Propaganda (0) & Delegitimize the Opponent (3) & 567 &
russia (15), ukraine (11), nazi (10), russian (10), war (10), propaganda (8), people (7), state (6), nazis (5), terrorist (5) \\

Distort Reality (1) & Delegitimize the Opponent (3) & 134 &
ukraine (12), russia (5), ukrainian (4), nazi (3), azov (3), battalion (3), weremember (3), neo (2), glorification (2), nazis (2) \\

Delegitimize (3) & Delegitimize (3) & 406 &
nazi (25), ukraine (20), war (7), nazis (6), ukrainian (6), propaganda (6), russian (5), russia (5), support (5), azov (4) \\
\bottomrule
\end{tabularx}
\caption{Illustrative excerpt of token frequency summaries for selected confusion-matrix cells (Qwen3; example file). Full token summary files for all models/settings are released in the project repository.}
\label{tab:qwen3-token-example}
\end{table*}
\section{Additional Confusion Matrices: Qwen3}
\label{app:qwen3-confusions}
\begin{figure}[ht]
    \centering
    \begin{subfigure}[t]{\columnwidth}
        \centering
        \includegraphics[width=0.50\linewidth]{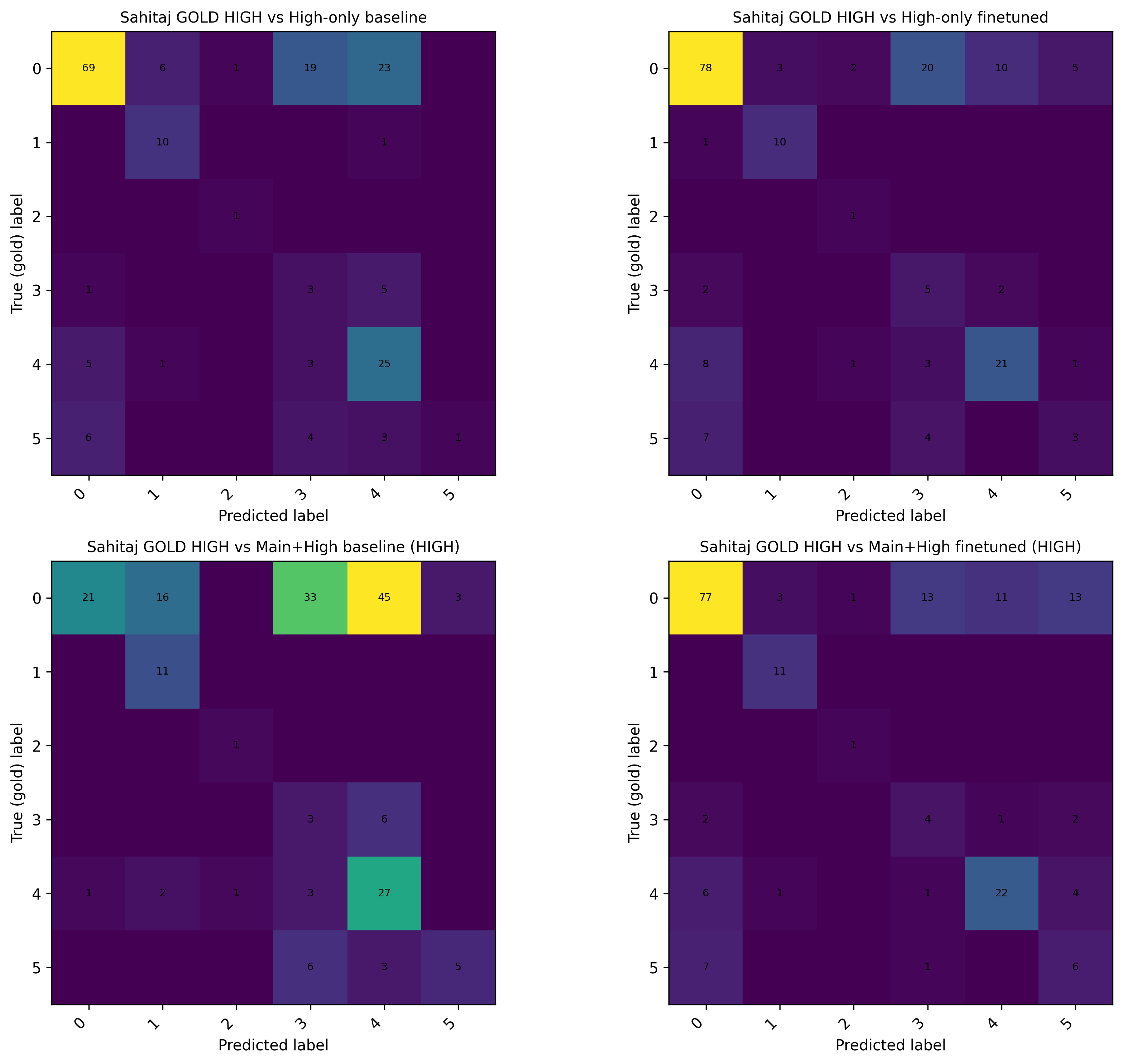}
        \caption{HIGH}
        \label{fig:qwen3-sahitaj-high}
    \end{subfigure}

    \vspace{0.5em}

    \begin{subfigure}[t]{\columnwidth}
        \centering
        \includegraphics[width=0.50\linewidth]{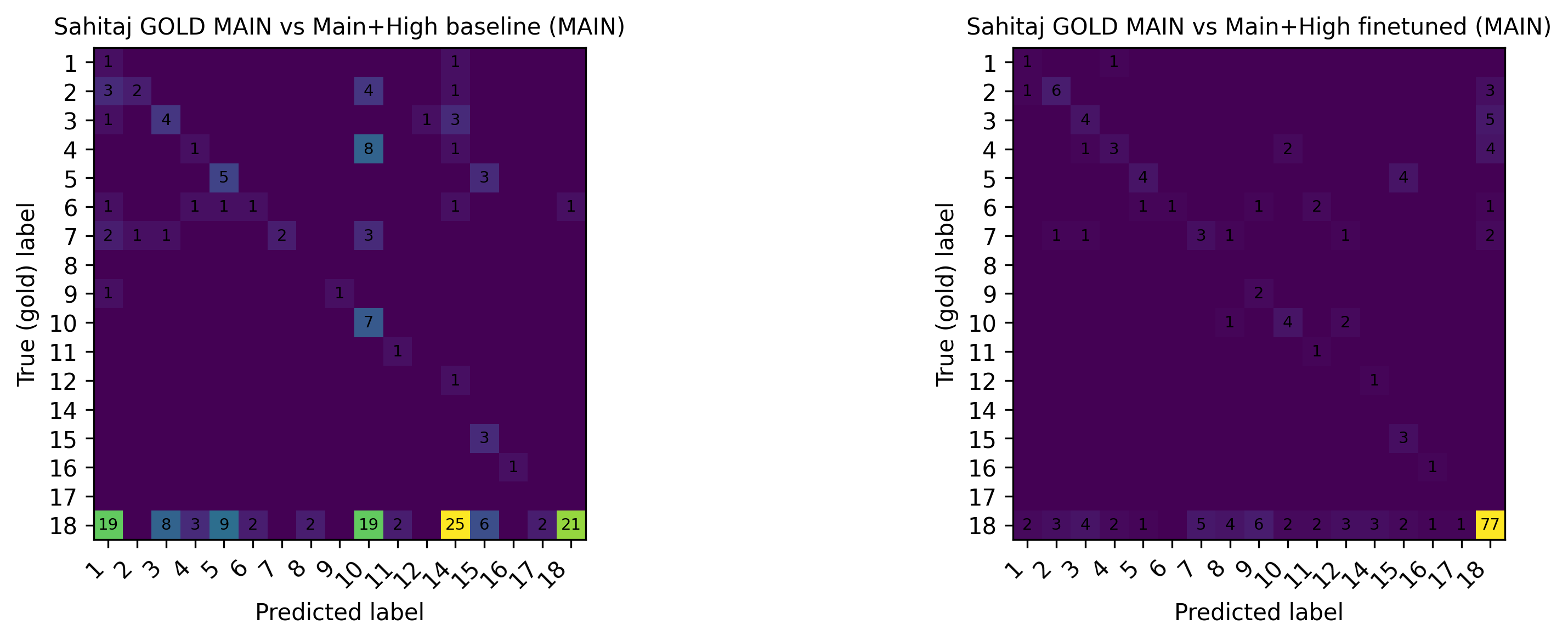}
        \caption{MAIN}
        \label{fig:qwen3-sahitaj-main}
    \end{subfigure}

    \caption{Qwen3 confusion matrices on the Sahitaj benchmark.}
    \label{fig:qwen3-sahitaj}
\end{figure}

\begin{figure}[ht]
    \centering
    \begin{subfigure}[t]{\columnwidth}
        \centering
        \includegraphics[width=0.50\linewidth]{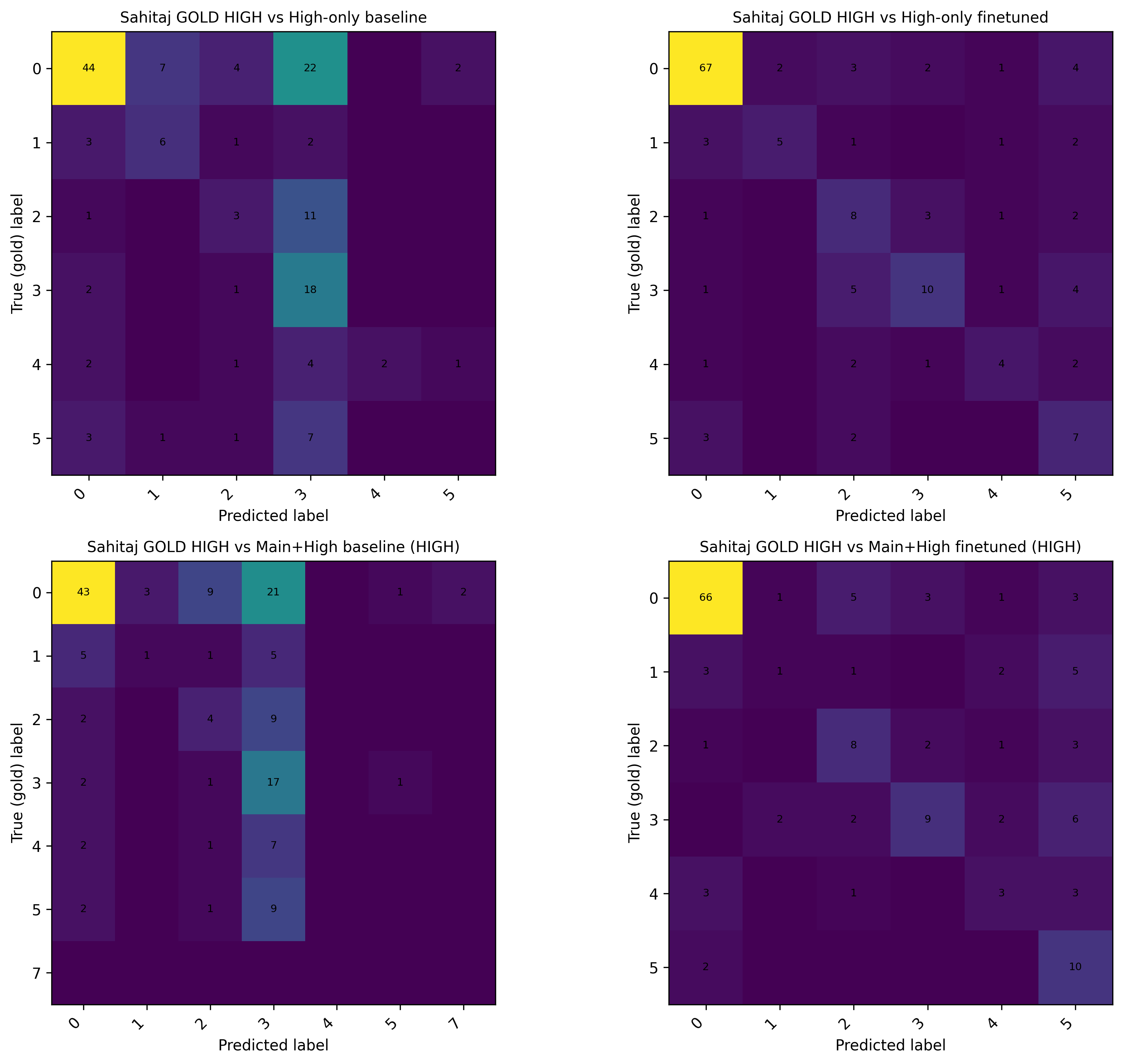}
        \caption{HIGH}
        \label{fig:qwen3-ours-high}
    \end{subfigure}

    \vspace{0.5em}

    \begin{subfigure}[t]{\columnwidth}
        \centering
        \includegraphics[width=0.50\linewidth]{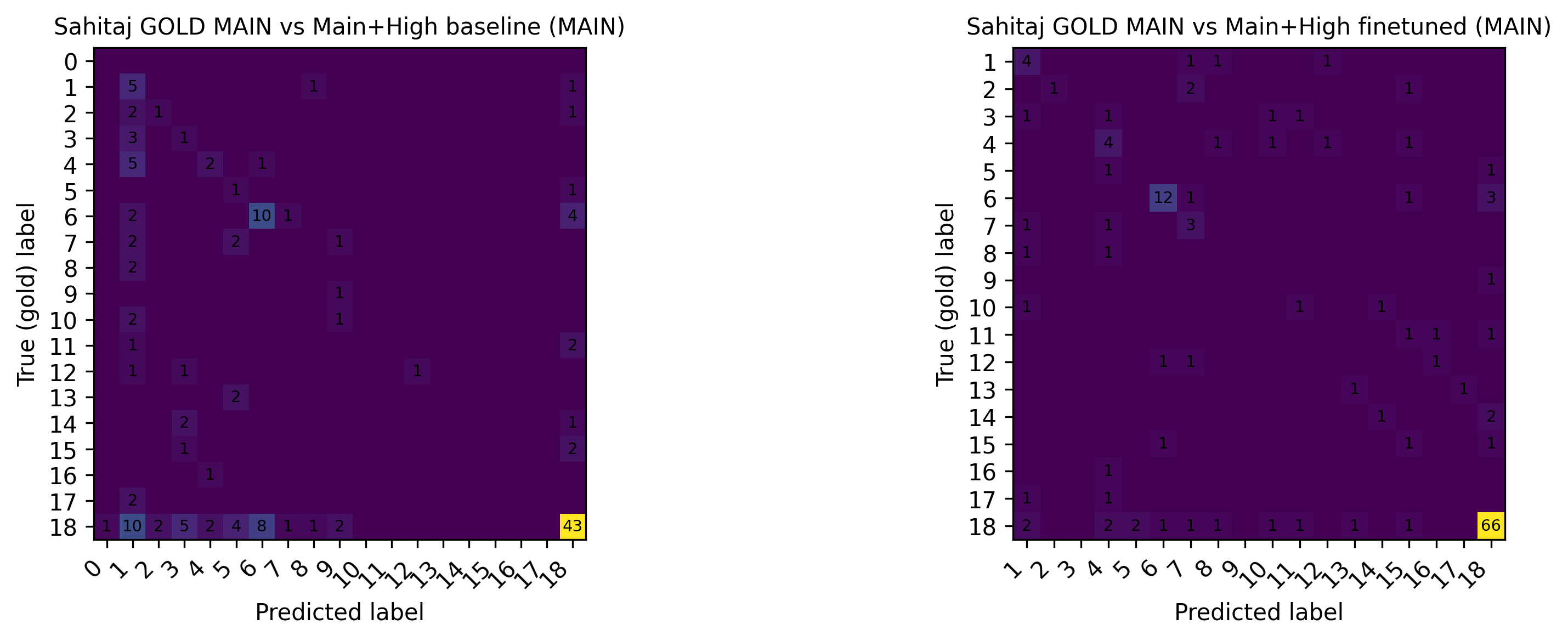}
        \caption{MAIN}
        \label{fig:qwen3-ours-main}
    \end{subfigure}

    \caption{Qwen3 confusion matrices on our dataset.}
    \label{fig:qwen3-ours}
\end{figure}
\end{document}